# Meta-Learning Guided Label Noise Distillation for Robust Signal Modulation Classification

Xiaoyang Hao, Zhixi Feng, *Member IEEE*, Tongqing Peng, and Shuyuan Yang*, *Senior Member IEEE*

*Abstract*—Automatic modulation classification (AMC) is an effective way to deal with physical layer threats of the internet of things (IoT). However, there is often label mislabeling in practice, which significantly impacts the performance and robustness of deep neural networks (DNNs). In this paper, we propose a meta-learning guided label noise distillation method for robust AMC. Specifically, a teacher-student heterogeneous network (TSHN) framework is proposed to distill and reuse label noise. Based on the idea that labels are representations, the teacher network with trusted meta-learning divides and conquers untrusted label samples and then guides the student network to learn better by reassessing and correcting labels. Furthermore, we propose a multi-view signal (MVS) method to further improve the performance of hard-to-classify categories with few-shot trusted label samples. Extensive experimental results show that our methods can significantly improve the performance and robustness of signal AMC in various and complex label noise scenarios, which is crucial for securing IoT applications.

*Index Terms*—Automatic modulation classification (AMC), label noise, meta-learning (ML), few-shot trusted label samples, multi-view signal (MVS).

## I. Introduction

WITH the rapid development of 5G technology and the Internet of Things (IoT), the electromagnetic environment is becoming more and more complex. The inherent mobile nature of most IoT devices increases their risk of exposure to untrusted electromagnetic environments. These devices are vulnerable to physical active attacks, such as deceptive jamming, pilot jamming, and sybil attacks [1], [2]. Automatic modulation classification (AMC) can be used to verify whether a signal is authorized or interfering. Therefore, AMC is an effective way to detect and identify physical layer threats, which is essential to ensure the security of communications and the reliability of IoT systems. In addition, AMC is of great significance for military and civilian applications such as dynamic spectrum access, jammer identification, and intruder detection [3]-[5].

Traditional AMC methods mainly include maximum likelihood methods [6] based on decision theory and pattern recognition methods [7] based on expert priors. These methods usually rely on expert features extracted from specific signal types. However, the expert features usually have problems such as unclear application scope, insufficient generalization, complex calculation, and difficult selection of parameters and types, which are difficult to meet IoT devices' robustness and real-time usage requirements. In recent years, AMC methods based on data-driven and deep learning have achieved great success, which can automatically extract high-level semantic features of signals. Convolution neural networks (CNNs) [8], [9], recurrent neural networks (RNNs) [10], [11], transformer networks (TNs) [12], generative adversarial networks (GANs) [13], [14], and their hybrid networks [15] achieve superior performance compared to traditional methods. However, the robustness of traditional methods and deep neural networks (DNNs) has been a key factor limiting their applications.

There are some studies to improve the robustness of AMC. In [16], an AMC method based on multi-task learning (MTL) was proposed, and its generalization ability was derived from multi-task feature learning with knowledge sharing in different noisy scenes. In [17], a joint framework consisting of two-channel spectral fusion, signal enhancement, and signal classification was proposed, which significantly improved the performance and robustness of signal recognition corrupted by channel noise by integrating a multi-level attention mechanism into the framework. In [18], the communication accumulation features were used to solve the severe AMC performance degradation caused by the multipath fading channel. Yun. L et al. explored the impact of adversarial examples on the robustness of DNN-based modulation recognition [19]. Furthermore, to improve the robustness of training and test data with distributional bias, K. Bu et al. proposed to perform the asymmetric mapping between datasets by adversarial training to improve domain generalization ability [20]. The above methods consider the signal sample noise caused by a complex electromagnetic environment. However, their performance also relies on an important assumption: a large amount of labeled data and that these labels are correct.

Like most tasks in computer vision [21], [22], the annotation of signals is expensive, time-consuming, and even error-prone. Solving the performance and robustness deterioration problem caused by signal labeling errors is the primary motivation of our study. We conclude that the signal labeling errors mainly come from the following aspects. 1) Difficult-to-describe signals cause label noise. For example, for scarce signals such as IoT attack against signals, non-cooperative communication signals, and new institutional signals, their description information and representative samples are often insufficient. 2) Indistinguishable signals cause label noise. For low signal-to-noise ratio signals, fine-grained signals (such as QAM16 and QAM64, or emitter signals with fingerprint differences), signals of various complex channels, and signals of different working states, it is difficult for even experts to judge the category. 3) Individual differences cause label noise. Differences in the professional level, subjective judgment, and working status of each worker will cause label noise. In addition, constant signals are usually segmented into multiple samples, and signal mislabeling can result in batch labeling errors for multiple signal samples. To sum up, it is difficult to guarantee the quality of signal labeling in practice. Therefore,

AMC is usually a weakly supervised learning (WSL) with inaccurate labels rather than strongly supervised learning. On the other hand, AMC is sensitive to label quality, and fitting to noisy samples will greatly reduce the generalization performance of the model or even bring disaster. For example, when noisy label samples are introduced as adversarial attack samples, there is a high probability of misjudging friends and foes. Also, label noise may increase the number of training features and model complexity. Therefore, weakly supervised AMC with label noise is crucial for IoT security, electronic countermeasures, and other applications.

To the best of our knowledge, signal label noise learning (SLNL) as a common inaccurate supervision has not been systematically studied. Depending on whether the labeling scenario involves experts, we roughly classify the existing label noise learning methods in computer vision into two categories: with trusted sample (WTS) reference and without trusted sample (OTS) reference. The main drawback of the WTS methods [23],[24] is that when the model falls into optimization bias by incorrectly correcting for noisy labels, this error can gradually accumulate and may lead to task failure. The main drawback of the OTS methods [25] is that they require a small amount of trusted labeled data as a guide. As we know from our practical experience, in most signal labeling scenarios, one or several experts lead many non-professional workers to complete the labeling task for several months or even longer, including completing the labeling task by themselves, outsourcing to a qualified team, or even crowdsourcing. Therefore, we can usually obtain a small amount of trusted labeled samples by experts and a large amount of untrusted labeled samples by non-professional workers. However, there are two key issues to be addressed for SLNL: 1) How to re-estimate the labels of untrusted samples. 2) How to extract, guide, and utilize a large number of untrusted labels with a small number of trusted labels.

To solve the above problems, we propose a teacher-student heterogeneous network (TSHN) framework. The teacher network is a meta-learning method for processing few-shot trusted labeled data, while the student network is a DNN for processing a large amount of untrusted labeled data. We re-evaluate the labels of untrusted samples with the idea that labels are representations. By comparing with the trusted data, we select the untrusted labeled samples with high confidence as purified data through the attention mask. Therefore, as the loss is continuously optimized, the purified data participates in the training of the student network in a self-training manner. In particular, for hard-to-classify categories with few trusted samples, we propose a multi-view signal (MVS) method to improve the guidance ability of the teacher network.

In summary, the main contributions of our work can be summarized as follows:
1) We propose TSHN for SLNL. Meta-learning based teacher network for trusted few-shot learning (FSL), while a DNN-based student network learns a large number of untrusted labeled samples with re-evaluated labels. To the best of our knowledge, this is the first work to study SLNL for robust AMC.
2) We re-evaluate labels by the idea of signal labels as feature representations. Through the joint loss optimization, the teacher network gradually guides untrusted labeled samples with higher confidence to participate in the learning of the student network. Untrusted labels with less confidence are distilled by an attention mask for further reuse.
3) A multi-view signal (MVS) method is proposed to further improve the performance of hard-to-classify categories with few trusted label samples and a large number of untrusted label noise samples.
4) For symmetric, asymmetric, and mixed label noise in practice, extensive experimental results show that our methods can significantly improve the performance and robustness of signal AMC.

The rest of this paper is organized as follows. The problem definition is presented in Section II. The proposed TSHN and MVS are introduced in Section III. The dataset construction method, experiments, and discussions are provided in Section IV. Lastly, the conclusions and future work are discussed in Section V.

## II. PROBLEM STATEMENT

Denote the modulated signal as the signal $r(t)$ can be written as:

$$r(t) = C(t)*s(t) + n(t) \qquad (1)$$

where $n(t)$ is additive white Gaussian noise (AWGN). $C(t)$ represents a response function of the complex radio channel. $*$ represents the time-domain convolution. The goal of AMC with label noise is to robustly identify the modulation type of $s(t)$ from $r(t)$. Without loss of generality, assume that the signal dataset with noisy labels is $D = \{(x_1, y_1),...,(x_n, y_n)\} \in (X,Y)_n$, and the unknown noise distribution is $P$, the goal of SLNL is to find the best mapping function $f: X \to Y$. The label noise model can be formulated as follows:

$$\tilde{y}_n = \begin{cases} i, i \in [k], i \neq y_n, prob \sum_{i \neq y_n} \eta_{x_n,i} = \eta_{x_n} \\ y_n, prob(1-\eta_{x_n}) \end{cases} \qquad (2)$$

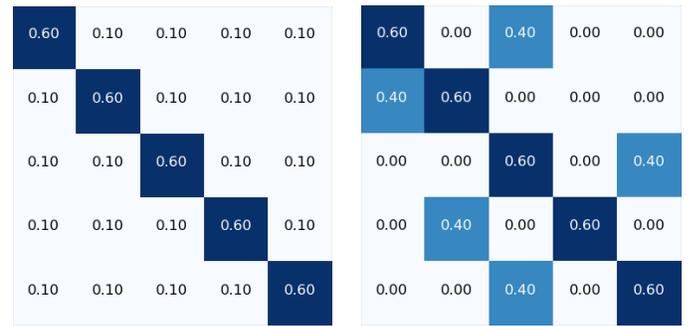

(a) SLN      (b) ALN (flip one)

Fig.1. Examples of SLN and ALN (flip one).

SLNL is mainly divided into symmetric label noise (SLN) and asymmetric label noise (ALN). In practice, it is possible that SLNL is a mixture of SLN and ALN. The generation process of SLN is completely random, and it can be understood that the real label $y_n$ is flipped to other label $i$ with the same probability $\eta_{x_n,i} = \eta$. ALN refers to the fact that real labels are flipped to other labels $i$ with a different probability $\eta_{x_n,i}$. For ALN, a certain class is more likely to be mislabeled as a specific label with a higher probability, $i \neq y_n$, $j \neq y_n$, $\eta_{x_n,i} > \eta_{x_n,j}$.

We measure the performance of the classifier through the loss function $L(f(x), y)$, and deal with label noise by minimizing empirical risk $R$. The empirical risk $R$ on signal dataset $D$ is defined as:

$$R_D(f) = \frac{1}{n}\sum_{i=1}^{n} L(f(x_i), y_i) \quad (3)$$

The result of empirical risk minimization is expressed as:

$$f^* = \arg\min_{f} R_D(f) \quad (4)$$

In practice, empirical risk minimization is insufficient to deal with diverse label noise. Therefore, it is necessary to combine the data itself, loss function, model structure, and training method for collaborative optimization.

## III. PROPOSED METHODS

In this section, we formally present TSHN and MVS for SLNL.

### A. Teacher-Student based Heterogeneous Network (TSHN) framework

Based on the idea that labels are representations, we calculate the soft labels of signal samples as their feature representations. The soft label is usually a vector with a sum of 1, which can describe how similar the signal sample is to each category. Therefore, we can re-evaluate the confidence of untrusted labeled samples by computing their soft labels. Specifically, for each untrusted sample, we compute its similarity to the class prototype of the trusted samples in the feature space, thereby obtaining a soft label. The samples with higher confidence are selected as purification-labeled samples, and the samples with lower confidence are distilled for further learning. During this process, few-shot trusted samples act as a guide to drive the meta-learning network to realize the evolution of features and prototypes. On the other hand, a large number of untrusted samples with high confidence are selected as purified samples to participate in the training of the student network. It is worth noting that the purified samples learn in a self-training manner. That is, as the training progress, the purified samples with greater confidence gradually participate in the correction of model parameters. The framework of TSHN is shown in Fig.2.

### 1) The Learning Process of Teacher Network

We improve the prototype network [26] as the teacher network, which is a meta-learning method for few-shot learning (FSL). During the meta-training process, the teacher network is trained on some meta-training tasks $\{\mathcal{T}_i\}_{i=1}^{G}$ sampled from a trusted set. The goal of meta-training is to learn a reliable model $m_\varphi$ based on the $J \times I$ supporting samples in a trusted support set ($J$ is the number of types and $I$ is the number of labeled samples in each class, and it is defined as a $J$-way $I$-shot meta-task.), which can classify the $J \times H$ unlabeled samples ($H$ is set to 15.) in the trusted query set into $J$ classes. By learning and extracting meta-knowledge for multiple meta-tasks, the teacher network can quickly transfer it to a new task with few-shot labeled samples. For the trusted support set and the trusted query set, we compute the cross-entropy loss between the predicted labels and the ground-truth labels.

$$L_t = -\frac{1}{U+V}\sum_{i=1}^{U+V}\sum_{c=1}^{N} y_i^c \log(F_\phi(z_i)) \quad (5)$$

where $F_\phi(z_i) = p(y_i | x_i)$ represents the predicted label obtained through the fully connected output. $U$ and $V$ represents the number of trusted support set samples and trusted query set samples sampled by the current episode from the trusted set, respectively.

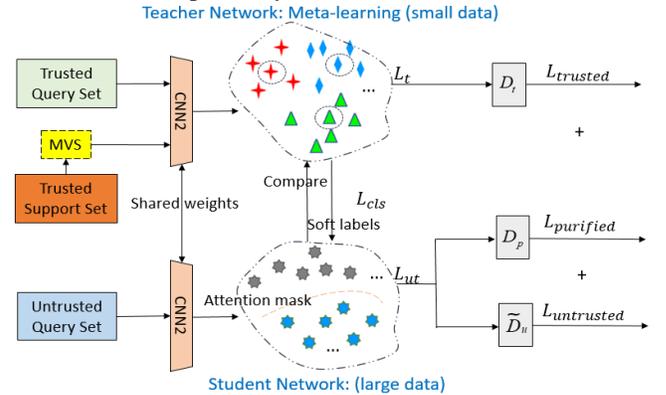

Fig.2. Teacher-Student Heterogeneous Network (TSHN) framework.

Different from the original prototype network, we start to calculate the prototype after the number of iterations, which helps get a good initial prototype. In addition, we update the prototype vector every five episodes with excessive changes in prototype vectors. Suppose there is an embedding space where each sample is clustered around its own class prototype. A sample $x$ is mapped to the feature space using a feature extraction network (CNN2 [8]) $F_\phi$ to obtain $F_\phi(x)$. Therefore, one can obtain the prototype $P_c$ of the $c$-th class by calculating the mean of all supporting samples of that class in the embedding space. The prototype of the $K$ support samples of the class $c$ is denoted as

$$P_c = \frac{1}{K}\sum_{y=c} F_\phi(x), \forall c = 1, ..., N \quad (6)$$

Denote the new prototype of class c calculated every $I_b$ iterations as $P_c^{new}$, then the prototype is updated as follows

$$P_c \leftarrow \xi P_c^{new} + (1-\xi) P_c \quad (7)$$

where $\xi$ represents the prototype update step size, for the relative stability of the prototype evolution process ($\xi \in (0,1)$ and $\xi = 0.3$ in this paper). Then the probability that the sample $x$ belonging to the class $c$ is

$$p(x^c) = \frac{\exp(-dist(F_\phi(x), P_c))}{\sum_{c'}^{N}\exp(-dist(F_\phi(x), P_{c'}))} \quad (8)$$

where $dist(\cdot,\cdot)$ can be some metrics such as Euclidean distance, cosine similarity, or a learnable metric. In this paper, we adopt the cosine similarity. $p(x^c)$ represents the probability that the untrusted labeled sample $x$ belongs to class $c$. Therefore, the soft label of $x$ can be expressed as

$$p(x) = [p(x^1), p(x^2), \ldots, p(x^N)] \quad (9)$$

The label confidence $c$ is expressed as

$$c \leftarrow \mu c + (1-\mu) p(x)^T y \quad (10)$$

where $\mu$ represents the confidence update step size, which is used for the relative stability of label confidence estimation results ($\mu \in (0,1)$ and $\mu = 0.6$ in this paper)

*2) The Learning Process of Student Network*

At the same time, we use CNN2 [8] as the student network, which shares the weights with the teacher network. For the untrusted query set, we compare it with the trusted support set for feature similarity to obtain soft labels and confidence scores. Its loss function is expressed as:

$$L_{ut} = -\sum_{i=1}^{W}\sum_{c=1}^{N} \eta_i y_i^c \log(F_\phi(z_i)) \quad (11)$$

where $W$ represents the number of samples sampled from the untrusted query set in the current episode. $\eta_i$ represents the mask operation, when the confidence of the sample is greater than $\delta$ ($\delta \in (0,1)$ and $\delta = 0.5$ in this paper), $\eta_i = c_i$, otherwise $\eta_i = 0$.

$$W' = \sum_{i=1}^{W}(\eta_i = c_i) \quad (12)$$

In summary, the loss on the training set is expressed as

$$L_{cls} = \frac{L_t + L_{ut}}{U + V + W'} \quad (13)$$

Through the guidance of the teacher network to the student network, the purified samples (untrusted samples with high confidence) gradually participate in the student network parameter learning, while the rest of the untrusted labeled samples are distilled out as noise samples. That is, the entire dataset is divided into three parts: a small number of trusted samples set $D_t$, a purified sample set $D_p$ (untrusted samples with higher confidence), and a noise sample set $D_u$ (untrusted samples with lower confidence). Although the TSHN distilled most of the untrusted label samples, we made further corrections with $D_t$. This process is similar to a teacher teaching a student to complete most of the problems with a small number of examples, and the problems that are difficult for the student to complete need to be reviewed and confirmed again. Based on the above insights, we use a divide-and-conquer strategy for further learning on $D_t$, $D_p$, and $D_u$. Specifically, for the trusted dataset $D_t$, we directly use the cross-entropy loss.

$$L_{trusted} = \sum_{i=1}^{|D_t|} \ell_{ce}(F_\varphi(x_i), y_i) \quad (14)$$

However, for purified data that may still have a small amount of label noise, directly using the cross-entropy loss will bring the following problems: 1) One-hot labels will make the network overconfident and may overfit to a small amount of label noise; 2) The one-hot label encourages the gap between the category and other categories to be as large as possible, and it is difficult to adapt to this situation due to the bounded gradient; 3) The loss of wrong label position is ignored, that is, the relationship between categories is not considered.

Label smoothing is a regularization technique that perturbs the target variable to make the model less certain about its predictions. That is, it limits the maximum probability of the softmax function so that the maximum probability is not much larger than the other labels (overconfidence). Therefore, for the purified dataset $D_p$, we design a cross-entropy loss with label smoothing to combat the small amount of label noise still present.

$$\ell_p = (1-\varepsilon)\ell_{ce} + \varepsilon \sum \frac{\ell_{ce}}{N} \quad (15)$$

where $\ell_{ce}$ is the standard cross-entropy loss, $\varepsilon$ is a small positive number ($\varepsilon \in (0,1)$ and $\varepsilon = 0.5$ in this paper), and $N$ is the number of classes. $\ell_p$ can be understood as: loss is a penalty for "predicted distribution and real distribution" and "predicted distribution and prior distribution (uniform distribution)".

$$L_{purified} = \sum_{i=1}^{|D_p|} \ell_p(F_\varphi(x_i), y_i) \quad (16)$$

For dataset $D_u$ which still has more noisy labels, we use a forward correction loss [27].

$$\ell_{corr}(p(y_i | x_i), y_i = e^k) = -\log \sum_{j=1}^{c} C_{jk} p(y_i^j | x_i) \quad (17)$$

where $e^k$ is a one-hot encoding form, representing the $k$th standard vector. $C \in \mathbb{R}^{c \times c}$ is a $N \times N$ noise transformation matrix that satisfies $C_{ij} = p(y = j | y = i)$. In this paper, $C_{ij}$ is estimated by the Gold Loss Correction (GLC) method [28] using a trusted dataset $D_t$. The loss function for the remaining noisy label data is

$$L_{untrusted} = \sum_{i=1}^{|D_u|} \ell_{corr}(F_\varphi(x_i), y_i) \quad (18)$$

In summary, the loss function in the training process of the entire neural network is as follows

$$L = \frac{L_{trusted} + L_{purified} + L_{untrusted}}{|D_t| + |D_p| + |D_u|} \quad (19)$$

*B. A Multi-View Signal (MVS) method*

Assuming that the time length of the signal sample is $t$, it is randomly divided into $N$ segments $s(t) = [s(t_1), s(t_2), \ldots, s(t_N)]$. Shuffle $s(t_1) \sim s(t_N)$ randomly and splicing the segments together in a new order, so that each shuffle can get a new view sample of the same type. For the waveform variation triggered at the splice, it can be interpreted as a phase interference.

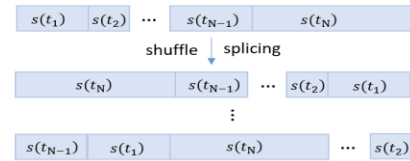

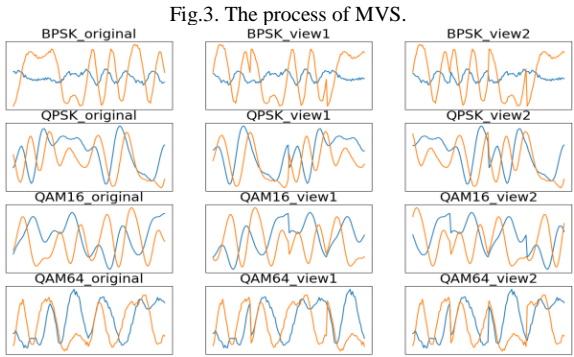

Fig.3. The process of MVS.

Fig.4. The example of MVS (N=4).

## IV. EXPERIMENTS AND DISCUSSION

### A. Datasets

We perform extensive experiments on two publicly available datasets with different channel environments and sample sizes.

**RadioML2016.10A [8]** Radio2016.10A dataset contains three analog modulations and eight digital modulations (from −20 dB to +18 dB, 2 dB apart SNR) that are widely used in IoT and wireless communications. The analog modulation modes are WB-FM, AM-SSB, and AM-DSB. The digital modulation modes are BPSK, QPSK, 8PSK, 16QAM, 64QAM, BFSK, CPFSK, and PAM4. The dataset contains a total of 220,000 samples. Each SNR contains 1000 samples for each modulation. Each sample includes in-Phase and quadrature (IQ) channels, and the data dimension is $2\times128$. The signal is modulated at a rate of roughly eight samples per symbol with a normalized average transmit power of 0dB. The dataset is generated in a harsh simulated propagation environment, including additive white Gaussian noise (AWGN), selective fading (Rician and Rayleigh), sample rate offset (SRO) and center frequency offset (CFO), etc.

**RadioML2016.04C [8]** The modulation modes are the same as the RadioML2016.10A dataset. This is a variable-SNR dataset with moderate LO drift, light fading, and numerous different labeled SNR increments for use in measuring performance across different signal and noise power scenarios. Furthermore, RadioML2016.04C contains fewer total samples (162,060) than RadioML2016.10A, which is used to validate the performance of SLNL with fewer label.

TABLE I
THE ACC (%) OF DIFFERENT METHODS WITH DIFFERENT LABEL NOISE RATES.

| Methods | Noise Ratio (RadioML2016.10A, SLN) | | | | | | | | | | |
|---|---|---|---|---|---|---|---|---|---|---|---|
| | 0.0 | 0.1 | 0.2 | 0.3 | 0.4 | 0.5 | 0.6 | 0.7 | 0.8 | 0.9 | 1.0 |
| MAE[29] | 38.23 | 38.19 | 37.01 | 36.26 | 34.69 | 31.38 | 27.63 | 15.67 | 7.88 | 8.56 | 7.19 |
| GCE[23] | 40.71 | 40.15 | 39.91 | 38.70 | 38.60 | 37.54 | 35.07 | 24.42 | 12.43 | 9.10 | 6.23 |
| GLC[29] | 48.32 | 47.82 | 47.74 | 46.91 | 46.57 | 45.13 | 45.08 | 44.16 | 43.61 | 41.22 | 39.79 |
| CNN2[8] | 49.64 | 48.95 | 48.47 | 47.69 | 46.25 | 44.71 | 39.30 | 26.44 | 14.63 | 9.09 | 9.09 |
| TSHN(Ours) | **50.90** | **50.21** | **49.82** | **49.01** | **49.00** | **48.73** | **48.21** | **47.88** | **47.25** | **46.33** | **45.93** |
| TSHN(↑) | 1.26 ↑ | 1.26 ↑ | 1.35 ↑ | 1.32 ↑ | 2.75 ↑ | 4.02 ↑ | 8.91 ↑ | 21.44 ↑ | 32.62 ↑ | 37.24 ↑ | 36.84 ↑ |

### B. Comparison methods and implementation details

We compare our method with MAE [29], GCE [23], GLC [29] and CNN2 [8], where MAE and GCE are the typical representatives of OTS methods and GLC is the typical representative of WTS methods. In addition, CNN2 with cross-entropy loss and two-layer convolution is a classical baseline approach for AMC. Both the comparison methods and our proposed TSHN use CNN2 as the feature extraction network, which facilitates fair comparison of all methods and references by later researchers. The hyper-parameters associated with the feature extraction network (CNN2) and the remaining hyper-parameters of the comparison methods use the values from the original paper. For our proposed TSHN, the teacher network uses the meta-task setting of "11way5shot". All algorithms are implemented with the PyTorch framework, and all experiments are conducted on NVIDIA RTX 3090 24G.

For the experimental setup, the training set, validation set, and test set are divided by 6:2:2. Then, the training set is further divided into two parts, the trusted set (default trusted fraction is 1%) and the untrusted set. For untrusted sets, experiments are performed on SLN, ALN, or a mixture of them. We use the noise rate to denote the proportion of noise labels to the untrusted labeled samples. In this paper, the noise rate takes 0 to 1, and the interval is 0.1. We repeat each experiment five times and average the accuracy to ensure the reliability of the results. More details will be introduced in the experimental part.

### C. Experiment 1: Investigation on the AMC with SLN

Based on the proposed method and the comparison methods, we first performed a comparative analysis of AMC with SLN. Based on the RadioML2016.10A dataset, the table I shows the average accuracy of the different methods with different label noise rates, where TSHN(↑) represents the accuracy improvement score compared to CNN2 at the corresponding noise rate. We can conclude that our proposed TSHN outperforms the baseline method CNN2 in all noise rate cases, and the performance and robustness improvement becomes more and more obvious as the noise rate increases. The maximum accuracy improvement reaches 37.24% when the noise rate is 0.9. The comparison of the curves in Fig.5(a) and Fig.5(d) more visually represents the variation of CNN2 and TSHN methods with noise rate (label noise) and signal-to-noise ratio (sample noise). The performance of the CNN2 method starts to seriously degrade until complete failure when the noise rate is greater than 0.5, while the TSHN method always maintains good performance and robustness. Moreover, for lower noise rates, CNN2 still exhibits a small amount of performance degradation. In addition, WTS methods (GLC and TSHN) significantly outperform CNN2 at label noise rates greater than 0.3, while OTS methods (GCE and MAE) show some failure or even inferiority to CNN2. The poor performance of OTS methods is perhaps due to its optimization bias affected by label noise, and the network is under-fitting. Fig.6 shows the classification confusion matrix of CNN2 and TSHN. It can be seen that TSHN greatly improves the classification performance due to the correction of label noise, while the remaining misclassification cases are mainly caused by sample similarity and sample noise. As shown in Fig.5 (b) and Fig.5 (e), our proposed method achieves similar results on RadioML2016.04C.

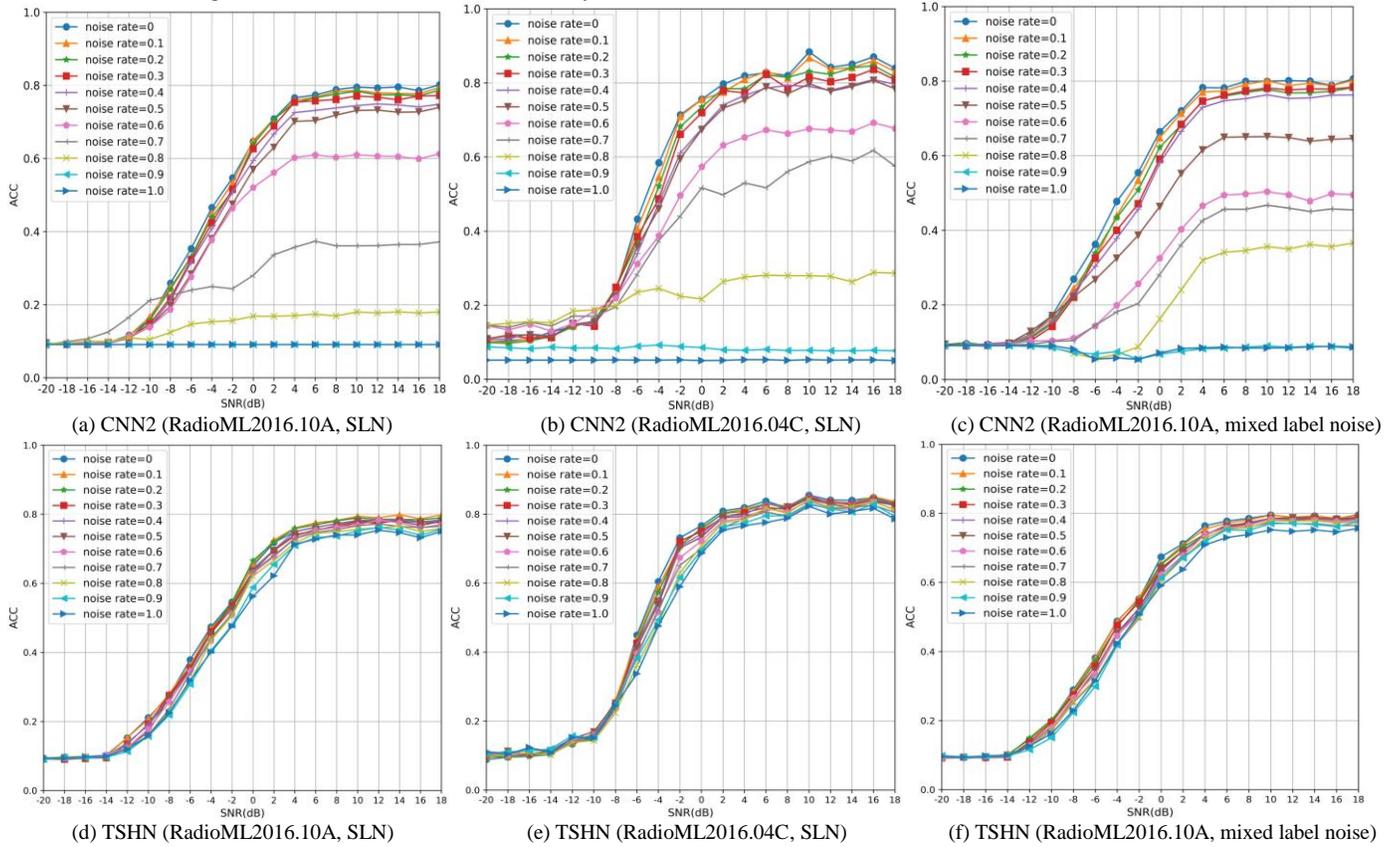

(a) CNN2 (RadioML2016.10A, SLN)    (b) CNN2 (RadioML2016.04C, SLN)    (c) CNN2 (RadioML2016.10A, mixed label noise)

(d) TSHN (RadioML2016.10A, SLN)    (e) TSHN (RadioML2016.04C, SLN)    (f) TSHN (RadioML2016.10A, mixed label noise)

Fig.5. Accuracy for different noise rates and SNRs.

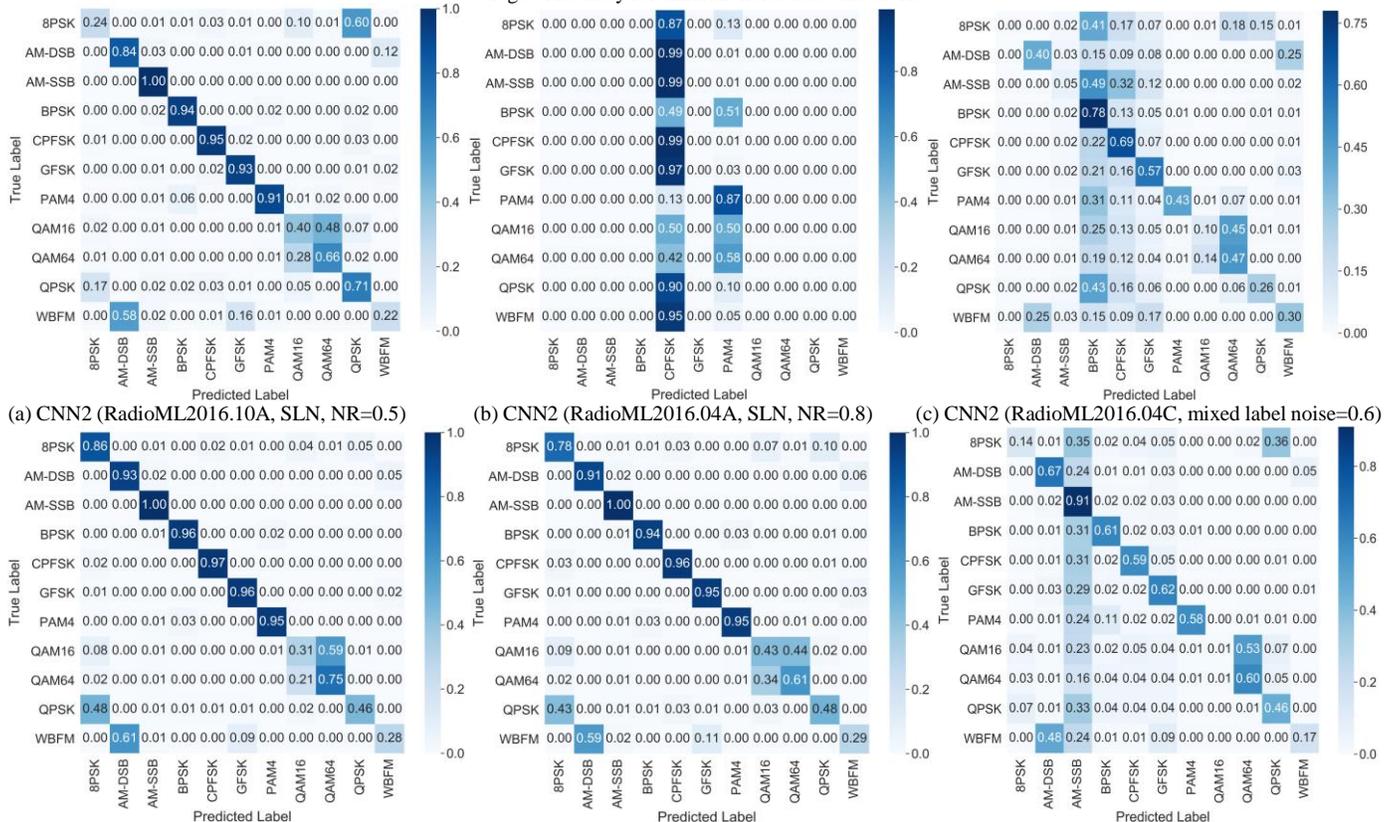

(a) CNN2 (RadioML2016.10A, SLN, NR=0.5)    (b) CNN2 (RadioML2016.04A, SLN, NR=0.8)    (c) CNN2 (RadioML2016.04C, mixed label noise=0.6)

(d) TSHN (RadioML2016.10A, SLN, NR=0.5)    (e) TSHN (RadioML2016.04A, SLN, NR=0.8)    (f) TSHN (RadioML2016.04C, mixed label noise=0.6)

Fig.6. Classification confusion matrix for CNN2 and TSHN (RadioML2016.10A uses non-negative SNR samples and RadioML2016.04C uses full SNR samples)

## D. Experiment 2: Investigation on the Few-Shot Trusted AMC with SLN

Although TSHN has shown good performance and robustness under 1% trusted fraction condition (about 1200 trusted labeled samples per class). However, only very few trusted labeled samples might be available under extreme conditions, such as the scarcity type at the tail end of the long tail distribution. In addition, scarcity types tend to have a high labeling noise rate. This is due to the fact that when there are very few reference samples, markers are more likely to mislabel samples of similar types as the scarce type. Therefore, we further investigate the performance of TSHN under the conditions of very few trusted samples and a higher label noise rate. The results in Table II show that the robustness improvement of TSHN becomes more and more significant as the number of trusted labeled samples gradually increase, and there is a substantial improvement at positive signal-to-noise ratios (with more label noise and fewer sample noise). Moreover, when the label noise rate is 0.8 and 1.0, TSHN with only 5 trusted labeled samples per class could achieve far better performance than CNN2. When there are 60 trusted labeled samples per class, TSHN outperforms CNN2 comprehensively. In conclusion, TSHN is still effective under the condition of few trusted labeled samples and more label noise, which can meet the robust AMC requirements of extreme IoT environments to a certain extent.

## E. Experiment 3: Investigation on the AMC with Mixed Label Noise

In practice, it is often a mixture of symmetric and asymmetric label noise due to factors such as the inherent indistinguishable properties of the sample, sample noise, or subjective staff factors. For this reason, we further explore the performance of TSHN. Specifically, we set ALN(flip one) for the indistinguishable categories in CNN2 [8] (QAM16 and QAM64, QPSK and 8PSK), while the rest of the categories are set to SLN. We call it mixed-label noise. It can be concluded from Fig.5(c) and Fig.5(f) that when the fraction ratio of the trusted labeled samples is 1% and the untrusted samples are mixed label noise, compared with CNN2, TSHN always maintains stronger robustness. TSHN(↑) in Table III shows the accuracy improvement of TSHN over CNN2 on RadioML2016.10A (Samples of all signal-to-noise ratios) and RadioML2016.04C (Samples of all signal-to-noise ratios), respectively. Among them, when the label noise rate is 0.8, and above, CNN2 completely fails, while TSHN(↑) reaches 28%~49%

## F. Experiment 4: Investigation on the Few-Shot trusted AMC with ALN (Flip One) and Difficulty Classes

In practice, indistinguishable categories tend to have a small number of trusted labeled samples and high label noise rates because it is difficult to determine the label. Therefore, in this section, we set up asymmetric label noise (QAM16 and QAM64, QPSK, and 8PSK) for hard-to-classify pairs and classify them with very few trusted labeled samples. Fig.7 shows the comparison results of CNN2, TSHN, and TSHN+ MVS. It can be concluded that TSHN improves the accuracy by 6.75%~9.03% compared with CNN2 under the experimental settings in this section. Moreover, our proposed MVS further improves the performance when used in TSHN, that is, the accuracy is improved by 13.55%~15.4% compared to CNN2. Among them, we perform MVC 20 times, and the multi-view signal samples are assigned the same labels as the original samples. According to the settings in this section, we can quickly obtain pseudo-labels of untrusted samples according to the learning of few-shot trusted AMC (TSHN+MVS) so as to assist and improve the labeling efficiency and accuracy of professionals.

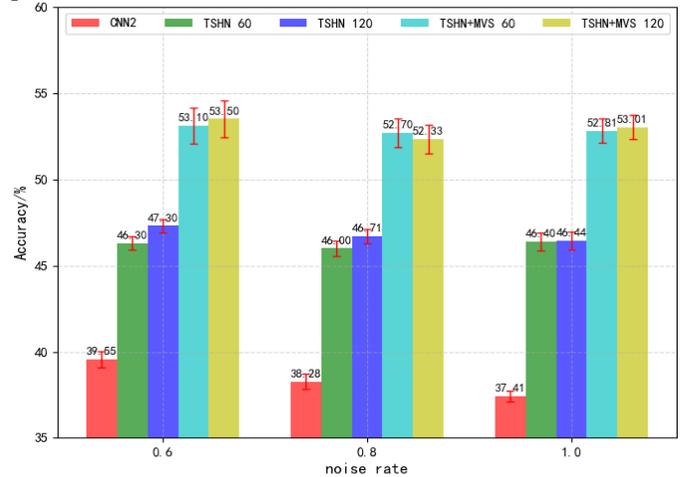

Fig.7. Accuracy of the Few-Shot trusted AMC with ALN (flip One) and difficulty classes (RadioML2016.10A, non-negative SNRs).

TABLE II
THE ACC (%) OF TSHN WITH DIFFERENT LABEL NOISE RATES AND VERY FEW TRUSTED LABELED SAMPLES.

| Noise Ratio | Trusted Samples | SNRs (RADIOML2016.10A, SLN) | | | | | | |
|---|---|---|---|---|---|---|---|---|
| | | -18 dB | -12 dB | -6 dB | 0 dB | 6 dB | 12 dB | 18 dB |
| 0.6 | 5 | 10.23 | 10.49 | 32.19 | 36.44 | 46.38 | 47.83 | 47.98 |
| | 10 | 10.40 | 10.41 | 31.28 | 45.23 | 56.12 | 60.56 | 60.29 |
| | 60 | 9.98 | 12.56 | 34.74 | 53.26 | 68.89 | 70.43 | 70.56 |
| | 120 | 10.00 | 10.95 | 33.50 | 55.59 | 71.18 | 73.64 | 73.78 |
| 0.8 | 5 | 10.42 | 10.35 | 26.66 | 31.49 | 38.56 | 40.23 | 40.99 |
| | 10 | 9.22 | 10.80 | 26.70 | 32.23 | 44.69 | 45.23 | 45.28 |
| | 60 | 10.01 | 11.28 | 30.24 | 42.14 | 62.25 | 63.20 | 63.21 |
| | 120 | 10.00 | 12.00 | 31.18 | 49.00 | 61.55 | 66.45 | 65.01 |
| 1.0 | 5 | 9.63 | 10.25 | 13.56 | 24.78 | 26.23 | 25.62 | 26.24 |
| | 10 | 10.47 | 9.63 | 22.46 | 27.27 | 31.29 | 33.71 | 31.26 |
| | 60 | 10.76 | 11.23 | 25.21 | 31.11 | 40.25 | 42.26 | 41.24 |
| | 120 | 10.00 | 11.09 | 29.77 | 36.42 | 43.09 | 44.59 | 44.18 |

First Author *et al.*: Title    8TABLE III
THE ACC (%) OF DIFFERENT METHODS WITH DIFFERENT LABEL NOISE RATES.

| Methods | Noise Ratio (RADIOML 2016.10A, mixed label noise) | | | | | | | | | | |
|---|---|---|---|---|---|---|---|---|---|---|---|
| | 0.0 | 0.1 | 0.2 | 0.3 | 0.4 | 0.5 | 0.6 | 0.7 | 0.8 | 0.9 | 1.0 |
| CNN2[8] | 50.38 | 49.36 | 47.99 | 47.45 | 46.47 | 40.18 | 29.72 | 28.19 | 20.05 | 8.10 | 8.08 |
| Ours | **50.50** | **50.37** | **49.72** | **49.33** | **48.97** | **49.00** | **48.56** | **47.60** | **47.92** | **47.36** | **46.67** |
| TSHN(↑) | 0.12 ↑ | 1.01 ↑ | 1.73 ↑ | 1.88 ↑ | 2.50 ↑ | 8.82 ↑ | 18.84 ↑ | 19.41 ↑ | 27.87 ↑ | 39.26 ↑ | 38.59 ↑ |
| Methods | Noise Ratio (RADIOML 2016.04C, mixed label noise) | | | | | | | | | | |
| | 0.0 | 0.1 | 0.2 | 0.3 | 0.4 | 0.5 | 0.6 | 0.7 | 0.8 | 0.9 | 1.0 |
| CNN2[8] | **55.26** | **54.80** | 53.46 | 53.28 | 51.55 | 50.35 | 44.81 | 40.45 | 9.08 | 2.63 | 2.89 |
| TSHN(ours) | 54.50 | 54.31 | **53.96** | **54.15** | **53.62** | **53.77** | **53.48** | **52.89** | **52.32** | **51.74** | **51.70** |
| TSHN(↑) | 0.76 ↓ | 0.49 ↓ | 0.50 ↑ | 0.87 ↑ | 2.07 ↑ | 3.42 ↑ | 8.67 ↑ | 12.44 ↑ | 43.24 ↑ | 49.11 ↑ | 48.81 ↑ |

## V. CONCLUSION

In this paper, we propose a robust AMC against label noise for the first time, which is derived from our practical application. To achieve this goal, we propose a meta-learning guided label noise distillation method called TSHN. TSHN combines the idea of the signal label as a representation to realize the divide and conquer of label noise samples. Furthermore, our proposed MVS further improves the label noise for hard-to-classify. Extensive experiments show that TSHN has a strong tolerance to all label noise rates. Moreover, TSHN shows better performance in the cases of symmetric, asymmetric, and mixed label noise. Even extreme cases perform well, including few-shot trusted labeled sample guide and hard-to-classify label noise. Therefore, our proposed methods are of great significance for ensuring applications such as IoT security and electronic countermeasures in complex and interference electromagnetic environments. In the future, we will explore robust AMC methods with better performance to deal with the dual problems of label noise and sample noise.

## REFERENCES

[1]  N. Wang *et al*., "Physical-layer security of 5G wireless networks for IoT: Challenges and opportunities," IEEE Internet Things J., vol. 6, no.5, pp. 8169-8181, 2019.
[2]  M. Serror *et al*., "Challenges and Opportunities in Securing the Industrial Internet of Things," IEEE Trans. Ind. Informat., vol. 17, no. 5, pp.2985-2996, 2021.
[3]  M. Liu *et al*., "Data-Driven Deep Learning for Signal Classification in Industrial Cognitive Radio Networks", IEEE Trans. Ind. Informat., vol. 17, no. 5, pp.3412-3421, 2021.
[4]  S. Yang *et al*., "One-Dimensional Deep Attention Convolution Network (ODACN) for Signals Classification", IEEE Access, vol. 8, pp. 2804–2812, 2020.
[5]  Q. Peng *et al*., "A Support Vector Machine Classification-Based Signal Detection Method in Ultrahigh-Frequency Radio Frequency Identification Systems", IEEE Trans. Ind. Informat., vol. 17, no. 7, pp.4646-4656, 2021.
[6]  S. Majhi *et al*., ''Hierarchical hypothesis and feature-based blind modulation classification for linearly modulated signals,'' IEEE Trans. Veh. Technol., vol. 66, no. 12, pp. 11057–11069, Dec. 2017.
[7]  A. Swami and B. M. Sadler, ''Hierarchical digital modulation classification using cumulants,'' IEEE Trans. Commun., vol. 48, no. 3, pp. 416–429,Mar. 2000.
[8]  T. J. O'Shea, T. Roy and T. C. Clancy, "Convolutional radio modulation recognition networks," in Proc. Int. Conf. Eng. Appl. Neural Netw., pp. 213–226, 2016.
[9]  T. J. O'Shea, T. Roy, and T. C. Clancy, "Over-the-air deep learning based radio signal classification," IEEE J. Sel. Topics Signal Process., vol. 12,no. 1, pp. 168–179, Feb. 2018.
[10] S. Rajendran, W. Meert, D. Giustiniano, V. Lenders and S. Pollin, ''Deep learning models for wireless signal classification with distributed low-cost spectrum sensors,'' IEEE Trans. Cognit. Commun. Netw., vol. 4, no. 3, pp. 433–445, Sep. 2018.
[11] Y. Chen, W. Shao, J. Liu, L. Yu and Z. Qian, "Automatic Modulation Classification Scheme Based on LSTM With Random Erasing and Attention Mechanism," in IEEE Access, vol. 8, pp. 154290-154300, 2020.
[12] J. Cai, F. Gan, X. Cao and W. Liu, "Signal Modulation Classification Based on the Transformer Network," IEEE Transactions on Cognitive Communications and Networking, vol. 18, no.9, pp. 1-9, 2022.
[13] K. Bu, Y. He, X. Jing and J. Han, "Adversarial Transfer Learning for Deep Learning Based Automatic Modulation Classification", IEEE Signal Processing Letters, vol. 27, pp. 880-884, 2020.
[14] Q. Wang, P. Du, X. Liu, J. Yang and G. Wang, "Adversarial unsupervised domain adaptation for cross scenario waveform recognition", Signal Processing, 2020.
[15] N. E. West and T. O'Shea, ''Deep architectures for modulation recognition,'' in Proc. IEEE Int. Symp. Dyn. Spectr. Access Netw. (DySPAN), pp. 1–6, Mar. 2017.
[16] Y. Wang *et al*., ''Multi-Task Learning for Generalized Automatic Modulation Classification Under Non-Gaussian Noise With Varying SNR Conditions,'' IEEE Trans. Wirel. Commun., vol. 20, no. 6, pp. 3587–3596,Mar. 2021.
[17] S. Lin, Y. Zeng and Y. Gong, ''Modulation Recognition Using Signal Enhancement and Multi-Stage Attention Mechanism,'' IEEE Trans. Wirel. Commun., vol. 20, no. 6, pp. 1-15,Mar. 2022.
[18] H.C. Wu, *et al*., ''Novel Automatic Modulation Classification Using Cumulant Features for Communications via Multipath Channels,'' IEEE Trans. Wirel. Commun., vol. 7, no. 8, pp. 3098–3105,Mar. 2008.
[19] Y. Lin, *et al*., ''Threats of Adversarial Attacks in DNN-Based Modulation Recognition,'' IEEE INFOCOM., July. 2020.
[20] K. Bu, Y. He, X. Jing and J. Han, "Adversarial Transfer Learning for Deep Learning Based Automatic Modulation Classification", IEEE Signal Processing Letters, vol. 27, pp. 880-884, 2020.
[21] C. Zhang, S. Bengio, M .Hardt, B. Recht, O. Vinyals, "Understanding deep learning (still) requires rethinking generalization", Communications of the ACM, vol64, pp 107–115, 2021.
[22] Y. Liu, N. Xu, Y. Zhang, X. Geng, "Label Distribution for Learning with Noisy Labels", International Joint Conference on Artificial Intelligence (IJCAI), pp. 2568-2574, 2020.
[23] Z. Zhang, M. Sabuncu, "Generalized Cross Entropy Loss for Training Deep Neural Networks with Noisy Labels", Conference on Neural Information Processing Systems (NeurIPS 2018), pp 1-14, 2018.
[24] J. Li, R. Socher, S. Hoi, "DivideMix: Learning with Noisy Labels as Semi-supervised Learning", International Conference on Learning Representations (ICLR), pp. 1-14, 2020.
[25] D. Hendrycks *et al*. "Using Trusted Data to Train Deep Networks on Labels Corrupted by Severe Noise", Advances in Neural Information Processing Systems (NeurIPS 2018), pp1-17, 2018.
[26] J. Snell, K. Swersky, and R. Zemel, "Prototypical Networks For Few Shot Learning," in Proc. Adv. Neural Inform. Process. Syst., 2017.
[27] G.Patrini *et al*., "Making deep neural networks robust to label noise: A loss correction approach.", IEEE Conference on Computer Vision and Pattern Recognition (CVPR), pp.1944–1952, 2017.
[28] B. Han *et al*., "Co-teaching: Robust training of deep neural networks with extremely noisy labels.", Conference and Workshop on Neural Information Processing Systems (NIPS), pp.8527–8537, 2018.
[29] A. Ghosh *et al*. "Co-teaching: Robust loss functions under label noise for deep neural networks.",(AAAI), pp.1919–1925, 2017.